\documentclass[10pt,twocolumn,letterpaper]{article}

\usepackage{iccv}
\usepackage{times}
\usepackage{epsfig}
\usepackage{graphicx}
\usepackage{amsmath}
\usepackage{amssymb}
\usepackage{amsbsy}
\usepackage{array}
\usepackage{booktabs}
\usepackage{txfonts}

\DeclareMathOperator*{\argmax}{arg\,max}
\usepackage[breaklinks=true,bookmarks=false]{hyperref}

\iccvfinalcopy % *** Uncomment this line for the final submission

\def\iccvPaperID{****} % *** Enter the ICCV Paper ID here

\ificcvfinal\fi

\begin{document}

%%%%%%%%% TITLE
\title{Continual Learning with Deep Streaming Regularized Discriminant Analysis}

\author{Joe Khawand $^{1,2}$ \and Peter Hanappe $^{2}$ \\ \\
$^1$ Ecole Polytechnique\\
$^2$ Sony Computer Science Laboratories Paris\\
%6 Rue Amyot, 75005 Paris\\
{\tt\small joe.khawand.20@polytechnique.org}
\and David Colliaux $^{2}$\\
%\and
}

\maketitle
\ificcvfinal\thispagestyle{empty}\fi

%%%%%%%%% ABSTRACT
\begin{abstract}
Continual learning is increasingly sought after in real-world machine learning applications, as it enables learning in a more human-like manner. Conventional machine learning approaches fail to achieve this, as incrementally updating the model with non-identically distributed data leads to catastrophic forgetting, where existing representations are overwritten. Although traditional continual learning methods have mostly focused on batch learning, which involves learning from large collections of labeled data sequentially, this approach is not well-suited for real-world applications where we would like new data to be integrated directly. This necessitates a paradigm shift towards streaming learning. In this paper, we propose\footnote{\url{https://github.com/SonyCSLParis/Deep_SRDA.git}} a streaming version of regularized discriminant analysis as a solution to this challenge. We combine our algorithm with a convolutional neural network and demonstrate that it outperforms both batch learning and existing streaming learning algorithms on the ImageNet ILSVRC-2012 dataset.
\end{abstract}

%%%%%%%%% BODY TEXT
\section{Introduction}
Continual learning, also known as lifelong learning, refers to the ability of a learning system to sequentially acquire and adapt knowledge over time. This type of learning mimics animal learning \cite{kudithipudi_biological_2022} and is increasingly sought after in various domains such as medical diagnostics \cite{lee_clinical_2020}, autonomous vehicles \cite{shaheen_continual_2022}, and finance \cite{philps_continual_2019}, where the learner needs to continually adapt to changing data. The major challenge in continual learning is the phenomenon of \textit{catastrophic forgetting} \cite{french_catastrophic_1999,mccloskey_catastrophic_1989}. It refers to the situation where a naively incrementally trained deep neural network forgets previously learned representations to specialise to the new task at hand.

Traditionally, the bulk of research \cite{van_de_ven_three_2022} in continual learning has primarily concentrated on batch learning approaches, which process data in fixed batches. In this setting, a continual learner typically iterates multiple times over the given task in an offline manner, allowing them to achieve satisfactory performance. However, this approach requires storing all data from the current task for training, which is not suitable for on-device learning.

As a result, recent research has emerged in the field of Online Continual learning \cite{mai_online_2022}, where data arrives in small, incremental batches and previously seen batches from the current or previous tasks are no longer accessible. Therefore, a model must effectively learn from a single pass over the online data stream, even when encountering new classes (Online Class Incremental, OCI) or data non-stationarity, such as new background, blur, noise, illumination, and occlusion (Online Domain Incremental, ODI).

\begin{figure}[t]
\begin{center}
   \includegraphics[width=0.84\linewidth]{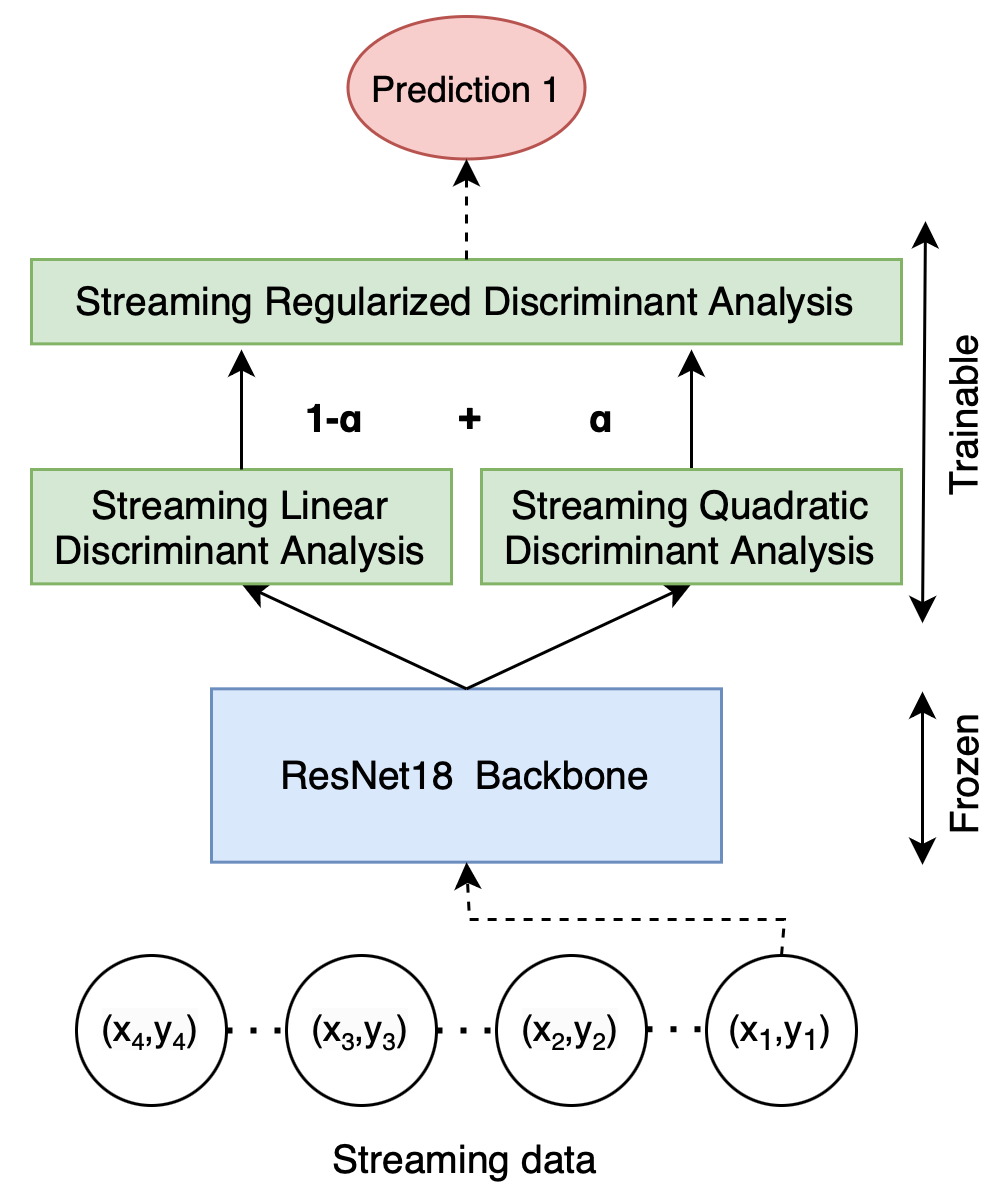}
\end{center}
   \caption{Deep SRDA model diagram.}
\label{fig:modelfig}
\end{figure}

We take this scenario one step further and consider a \textbf{streaming case of the online continual learning} scenario, where a learner learns from batches of size 1. This particular case of Streaming learning aims to develop methodologies that can efficiently learn from streaming data, enabling continuous adaptation without relying on complete batches. Specifically, we concentrate on the application of classification in computer vision, focusing on the general scenario of Online Class Incremental.

In this paper, we aim to contribute to the field of continual learning by proposing a novel approach called Deep Streaming Regularized Discriminant Analysis (SRDA). Building upon the foundations of Deep Streaming Linear Discriminant Analysis (SLDA) \cite{hayes_lifelong_2020}, our method combines  SLDA  with a streaming version of Quadratic Discriminant Analysis (QDA) to achieve state-of-the-art performance. As done in \cite{hayes_lifelong_2020}, we combine our model with a convolutional neural network (CNN) and empirically demonstrate its superiority over the other streaming and batch learning methods on the ImageNet ILSVRC-2012 dataset \cite{russakovsky_imagenet_2015}. To the best of our knowledge, this use of a regularized discriminant analysis method with a neural network represents a novel contribution that has not been explored before.

\textbf{This paper makes the following contributions:}
\begin{enumerate}
    \item We present the SQDA and SRDA algorithms. We show that SQDA does not generalize to high dimensional problems and present SRDA as a solution. %We are the first to use this approach for classification of a latent space of a CNN.
    \item We demonstrate that SRDA outperforms state-of-the-art streaming and continual learning algorithms.
\end{enumerate}

%------------------------------------------------------------------------
\section{Related work}
\subsection{Continual Learning}
Continual learning addresses the challenges of catastrophic forgetting that occur when training a model incrementally, breaking the usual i.i.d. assumption on the training data. This problem arises from the plasticity dilemma \cite{abraham_memory_2005} and has been heavily studied in recent years \cite{van_de_ven_three_2022}. Various continual learning scenarios have been developed to continually train models, with the three main ones being task incremental, class incremental, and domain incremental.

In the task incremental setting, different training steps are identified by a label. Models trained in this setting tend to perform well because there is an indication of the task in the data. However, this scenario is not very realistic as real-world data is not typically labelled by tasks. The class incremental scenario considers the real-world scenario of adding new classes without any task delimiters. Lastly, in the domain incremental scenario, the focus is on dealing with the addition of new domains or changing environments without any explicit task labels. 

To mitigate catastrophic forgetting, several methods have been employed. The main ones include regularization \cite{li_learning_2017,triki_encoder_2017,kirkpatrick_overcoming_2017,zenke_continual_2017,aljundi_memory_2018,lee_overcoming_2018}, rehearsal or pseudo-rehearsal \cite{rolnick_experience_2019,shin_continual_2017,rios_closed-loop_2020}, combined \cite{kemker_fearnet_2018,buzzega_dark_2020,li_learning_2022}, and architectural \cite{mallya_packnet_2018,li_learn_2019}. Notably, the rehearsal and pseudo-rehearsal categories have shown the most promising results. In these approaches, the learner stores previously encountered samples in a buffer for future training \cite{rolnick_experience_2019}. In some instances, pseudo-rehearsal techniques explore the replacement of the buffer with a generative model \cite{shin_continual_2017,rios_closed-loop_2020}.
\subsection{Online Continual Learning}
\label{sec:ocl}
Online continual learning is a more challenging subset of continual learning where data arrives in an online fashion one tiny batch at a time and previously encountered batches are not accessible. This field builds upon existing methods for continual learning while adding specific tricks to tackle this scenario. In this online setting, recent works \cite{masana_class-incremental_2022,hou_learning_2019,wu_large_2019,ahn_ss-il_2022,mai_online_2022} have shown that the Softmax layer and its associated Fully-Connected layer suffer from \textit{task-recency bias}, where those layers tend to be biased to the last encountered classes. This has prompted the creation of multiple tricks to alleviate this problem. One example is the application of various tricks in replay-based scenarios:

\begin{itemize}
    \item \textbf{Labels Trick} \cite{zeno_task_2019}: Cross-entropy loss calculation considers only the classes present in the mini-batch, preventing excessive penalization of logits for classes absent from the mini-batch.
    \item \textbf{Multiple Iterations} \cite{aljundi_online_2019}: A single mini-batch is stored in a buffer and iterated upon multiple times. In addition to that, previously stored experiments are also replayed.
    \item \textbf{Nearest Class Mean Classifier}: Replaces the last biased fully connected classification layer by a nearest mean classifier such as in iCarl \cite{rebuffi_icarl_2017}.
    \item \textbf{Separated Softmax} \cite{ahn_ss-il_2022}: Since one softmax layer results in a bias explained in \cite{masana_class-incremental_2022,hou_learning_2019,wu_large_2019,ahn_ss-il_2022,mai_online_2022}, this technique employs two Softmax layers one for old classes and one for new classes. Thus training new classes will not overly penalize the old logits.
    \item \textbf{Review trick} \cite{ferrari_end--end_2018}: Adds an additional fine-tuning step using a balanced subset of the memory buffer. This trick is used in the End-to-End method used in our benchmarks \ref{table1}.
\end{itemize}

However, when the batch size is reduced to one, the Stochastic Gradient Descent (SGD) usually employed in most of these methods becomes noisy, making convergence challenging. This is precisely where streaming learning comes into play.

\subsection{Streaming Learning}
Streaming learning, a field of study since 1980 \cite{mwnro_selection_nodate}, primarily focuses on i.i.d data streams and utilizes online learning methods. However, due to the Softmax bias and SGD instability for batches of size 1, regular online learning methods are not optimal for streaming learning on \textbf{non-i.i.d data streams}.

For this case, one area of streaming learning considers the use of streaming decision trees \cite{oliver_streaming_2021}, where Hoeffding decision trees \cite{hulten_mining_2001} are adapted to avoid catastrophic forgetting. Those can also be combined into Streaming forests \cite{oliver_streaming_2021,joshua_lifelong_2023}. However, the issue with these types of methods is that they are slow to train \cite{gaber_survey_2007}, and require extensive hyperparameter tuning, making them unsuited for fast-paced streaming scenarios and real-time on-device learning. 

Another approach involves employing a Nearest Mean Classifier \cite{mensink_distance-based_2013} instead of a Softmax layer. Research conducted by \cite{mai_supervised_2021} has demonstrated that this simple yet effective substitute not only addresses recency bias but also avoids structural changes in the Fully-Connected layer when new classes are encountered. Notably, this method has been effectively employed by iCarl \cite{rebuffi_icarl_2017}.

Another method used is Exstream \cite{hayes_memory_2019}. This method only updates fully connected layers of a CNN while maintaining a prototype for each class. It also has a policy for managing the buffer when it is full, merging the two closest exemplars. But as we will see in section \ref{sec:computation}, this method suffers in terms of computation time as it requires, in this case, 64 hours to run on our experiment, whereas SRDA requires 12 hours.

Especially relevant to this paper is Streaming LDA \cite{pang_incremental_2005} that was first used for data streams and has since been adapted in \cite{hayes_lifelong_2020} to work with CNNs. SLDA uses running class means and a common covariance matrix for all classes to assign labels to inputs based on the closest Gaussian distribution.

%------------------------------------------------------------------------
\section{Problem Setting}
We consider ensembles $\mathcal{X}$ and $\mathcal{Y}$, representing our datapoints and labels, respectively. We aim to train a model $F$ with parameters $\theta$ to accurately classify classes in  $\llbracket1, C\rrbracket$, where $C \in \mathbb{N}^{*}$. To achieve this, we adopt a streaming fashion approach, where each datapoint $x\in \mathcal{X}$ is individually sent to the model for fitting. Additionally, we adopt a class incremental scenario by ordering the samples in batches of classes. We consider this type of scenario to be the most general as it is similar to animal and human learning scenarios. 
%------------------------------------------------------------------------
\section{Deep Streaming RDA}
\label{sec:srda}
%Formally as Hayes and Kanan presented \cite{hayes_lifelong_2020}, we divide our model into a composition of two distinct functions $G$ and $F$. $G$ consists of the first few layers of a CNN (here a resnet18\cite{he_deep_2015}) and $F$ consists of our SRDA head. Because the filters learned in the early layers of a CNN vary little across large natural image datasets and are highly transferable \cite{yosinski_how_2014}, we chose to freeze the representation of $G$ and to only train $F$.

Similar to Hayes and Kanan's work \cite{hayes_lifelong_2020}, our model can be formally divided into a composition of two distinct functions $G$ and $F$, such as $y=F(G(x))$. $G$ is comprised of the initial layers of a CNN, specifically a ResNet-18 \cite{he_deep_2015} in our case, while $F$ represents our SRDA head. The early layers of a CNN, such as those in $G$, tend to learn filters that exhibit minimal variation across large natural image datasets and demonstrate high transferability \cite{yosinski_how_2014}. Therefore, we made the decision to freeze the parameters of $G$ and solely focus on training $F$.

The following subsections will present our SRDA model. We will start by presenting discriminant analysis before presenting an initial quadratic streaming version that led to our deep SRDA algorithm.

\subsection{Discriminant Analysis}
Discriminant analysis is a traditional machine learning algorithm that can be used for classification \cite{hastie_elements_2009}. It works on the hypothesis that the data follows a Gaussian multivariate distribution that is used to calculate the log posterior probability using Bayes' rule.
\medbreak

\noindent For each training example $x\in \mathcal{X}$ and $k\in \llbracket1,C\rrbracket$, the goal is to calculate the posterior probability in order to classify correctly. The Bayes' rule on the posterior probability of being in class $k$ for an element $x$ is:
\begin{equation}
    P(y=k|\boldsymbol{x})=\frac{P(\boldsymbol{x}|y=k) P(y=k)}{P(\boldsymbol{x})}
\label{eq:bayesrule}
\end{equation}

\noindent With $P(\boldsymbol{x}|y=k)$ modeled as a multivariate Gaussian distribution with a mean $\mu_{k}$ and a covariance $\Sigma_{k}$:
\begin{equation}
    P(\boldsymbol{x}|y=k)=\frac{\exp{(-\frac{1}{2}(\boldsymbol{x}-\boldsymbol{\mu}_{k})^{t}\boldsymbol{\Sigma}_{k}^{-1}(\boldsymbol{x}-\boldsymbol{\mu}_{k})})}{(2\pi)^{C/2}|\boldsymbol{\Sigma}|^{1/2}}
\label{eq:density}
\end{equation}

\noindent According to equations \ref{eq:bayesrule} and \ref{eq:density}, the log of the posterior or \textbf{discriminant $\gamma_{k}$} is given as follows:
\begin{equation}
\begin{split}
    \gamma_{k}=-\frac{1}{2}\log{|\boldsymbol{\Sigma}_{k}|}-\frac{1}{2}(\boldsymbol{x}-\boldsymbol{\mu}_{k})^{t}\boldsymbol{\Sigma}_{k}^{-1}(\boldsymbol{x}-\boldsymbol{\mu}_{k})\\ +\log{P(y=k)}+B
\end{split}
\label{eq:gammadisc}
\end{equation}

\noindent Where $B \in \mathbb{R}$ is a constant.

\noindent Finally, the classification rule is written as:
\begin{equation}
    F(\boldsymbol{x})=\argmax_{k} \gamma_{k}
\end{equation}

 With no further assumptions, this is referred to as Quadratic Discriminant Analysis (QDA). Linear Discriminant Analysis (LDA) and the streaming version of it \cite{hayes_lifelong_2020}, constrains equation \ref{eq:gammadisc} and considers equal covariance matrices between classes.

\subsection{Streaming Discriminant Analysis}
\subsubsection{Quadratic}
\label{sec:quadratic}
In order to adapt equation \ref{eq:gammadisc} to streams of data we need to calculate $\mu_{k}$, $\Sigma_{k}$, and $\Sigma_{k}^{-1}$ in a streaming fashion. We choose to replace those values by their \textbf{empirical estimators}. 

We consider a new element $z_{t}$ and $k\in\llbracket1,C\rrbracket$. The update functions are written as follows, where:
\begin{itemize}
    \item $c$ is the vector of encountered classes:
    \begin{equation}
        c_{(k=y,t+1)}=c_{(k=y,t)}+1
    \end{equation}
    \item $\hat{\mu}$ is the saved class means:
    \begin{equation}
        \boldsymbol{\hat{\mu}}_{(k=y,t+1)}=\frac{c_{(k=y,t)}\boldsymbol{\hat{\mu}}_{(k=y,t)}+\boldsymbol{z}_{t}}{c_{(k=y,t)}+1}
    \end{equation}
    \item $\hat{\Sigma}$ is the vector containing all the class covariance matrices:
    \begin{equation}
    \label{eq:sigma_qda}
        \boldsymbol{\hat{\Sigma}}_{(k,t+1)}=\frac{t\boldsymbol{\hat{\Sigma}}_{(k,t)}+\boldsymbol{\Delta_{t}}}{t+1}
    \end{equation}
    \begin{equation}
        \boldsymbol{\Delta_{t}}=\frac{t(\boldsymbol{z}_{t}-\boldsymbol{\hat{\mu}}_{(k=y,t)})(\boldsymbol{z}_{t}-\boldsymbol{\hat{\mu}}_{(k=y,t)})^{T}}{t+1}
    \end{equation}
    \item $\Lambda$ is the inverse of $\Sigma$ regularized with a shrinkage coefficient $\epsilon$:
    \begin{equation}
    \label{eq:lamda}
        \boldsymbol{\Lambda}_{(k,t)}=[(1-\epsilon)\boldsymbol{\hat{\Sigma}}_{(k,t)}+\epsilon \boldsymbol{I}]^{-1}
    \end{equation}
    \item $P(y=k)$ is calculated by incrementally and uniformly updating it for seen classes at time $t$. For a balanced dataset, this factor can be considered constant but is important for unbalanced datasets serving as a corrective term. %not so sure about it
    \begin{equation}
    \label{eq:prior}
        P(y=k)_{t}=\frac{c_{(k=y,t)}}{\sum_{n=1}^{C} c_{(k=n,t) }}
    \end{equation}
\end{itemize}

%\begin{equation}
%    \Sigma_{t+1}=\frac{t\Sigma_{t}+\Delta_{t}}{t+1}
%\end{equation}

Applying those updates to equation \ref{eq:gammadisc} leads to a streaming version of QDA mentioned by Hayes and Kanan \cite{hayes_lifelong_2020}. But the problem with SQDA is that, in high dimensionality, the number of datapoints needed to correctly empirically estimate the covariance matrices of each class are high \cite{friedman_regularized_1989}. As we will show in section \ref{sec:Experiments}, this approach struggles to translate to our high dimensional problem and mostly works with \textbf{low dimensional} datasets or ones with numerous examples per class. \textbf{This prompted us to look for a regularized alternative that solves this issue.}

\subsubsection{Regularized}

Friedman \cite{friedman_regularized_1989} proposed a compromise between LDA and QDA, that shrinks the separate covariances of QDA toward a common covariance as in LDA. Using a coefficient $\alpha\in[0,1]$, the regularization targets the class covariance matrices as follows:

\begin{equation}
    \boldsymbol{\Bar{\Sigma}}_{(k,t)}=\alpha \boldsymbol{\hat{\Sigma}}_{(k,t)} +(1-\alpha) \boldsymbol{\hat{\Sigma}}_{t'}
\end{equation}

 Where $\hat{\Sigma}_{(k,t)}$ is the empirical class covariance calculated through QDA (eq.\ref{eq:sigma_qda}), and $\hat{\Sigma}_{t'}$ the empirical covariance matrix calculated with SLDA, in this equation \ref{eq:sigma_lda}:

 \begin{equation}
 \label{eq:sigma_lda}
        \boldsymbol{\hat{\Sigma}}_{t'+1}=\frac{t'\boldsymbol{\hat{\Sigma}}_{t'}+\boldsymbol{\Delta}_{t'}}{t'+1}
 \end{equation}

\noindent Replacing this new regularised $\Bar{\Sigma}$ in equations \ref{eq:lamda} and \ref{eq:gammadisc} gives us \textbf{SRDA.}

The coefficient $\alpha$ controls the degree of shrinkage of the individual class covariance matrix estimates towards the pooled estimate. Since it is often the case that even small amounts of regularization can largely eliminate quite drastic instability \cite{titterington_statistical_1985}, some values of $\alpha$ have the potential of superior performance when the population class covariances substantially differ \cite{friedman_regularized_1989}. This performance boost is clearly shown in section \ref{sec:Experiments}.

%\begin{algorithm}
%\caption{Training Deep SRDA}\label{alg:cap}
%\textbf{Input:} $(X,Y)$ a set of data points and labels\\
%$n \in \mathbb{N}^{*}$, number of points in dataset.\\
%$d \in \mathbb{N}^{*}$, dimension of each vector $x \in X$.\\
%$C\in \mathbb{N}^{*}$, number of classes.\\
%$F$, pretrained DNN.
%\begin{algorithmic}[1]
%\State $\mu_{k} \gets 0_{d}, \;\; \forall k \in \llbracket 1,C \rrbracket$
%\State $\Sigma_{k} \gets 1_{d\times d}, \;\; \forall k \in \llbracket 0,C \rrbracket$
%\State $\Sigma \gets 1_{d\times d}$
%\State $\Lambda_{k} \gets 1_{d\times d}, \;\; \forall k \in \llbracket 0,C \rrbracket$
%\State $\Lambda \gets 1_{d\times d}, \;\; \forall k \in \llbracket 0,C \rrbracket$
%\For {$(x,y) \in (X,Y)$}
%\If{$N$ is even}
%    \State $X \gets X \times X$
%    \State $N \gets \frac{N}{2}$  \Comment{This is a comment}
%\ElsIf{$N$ is odd}
%    \State $y \gets y \times X$
%    \State $N \gets N - 1$
%\EndIf
%\EndFor
%
%\end{algorithmic}
%\end{algorithm}

%------------------------------------------------------------------------
\section{Experiments \& Results}
\label{sec:Experiments}

\subsection{Baselines}
We conducted a comprehensive analysis by comparing our method with both streaming methods and batch streaming methods. To evaluate the performance of our model, we utilized the metric described in \cite{hayes_memory_2019,kemker_measuring_2018}, which involves normalizing a model's performance by the offline model's performance: 
\begin{equation}
\label{eq:omega}
    \Omega_{all}=\frac{1}{T}\sum_{t=1}^{T}\frac{\rho_{t}}{\rho_{\textit{offline},t}}
\end{equation}
In our case, $\rho_{t}$ refers to the top-5 accuracy of our model at time $t$.

An $\Omega_{all}$ of 1 indicates that the continual learner performs equally well as the offline model. While it is theoretically possible to achieve results higher than one if the continual learner outperforms the offline model, such instances are rare in practice.

\begin{figure}[b]
\begin{center}
   \includegraphics[width=0.95\linewidth]{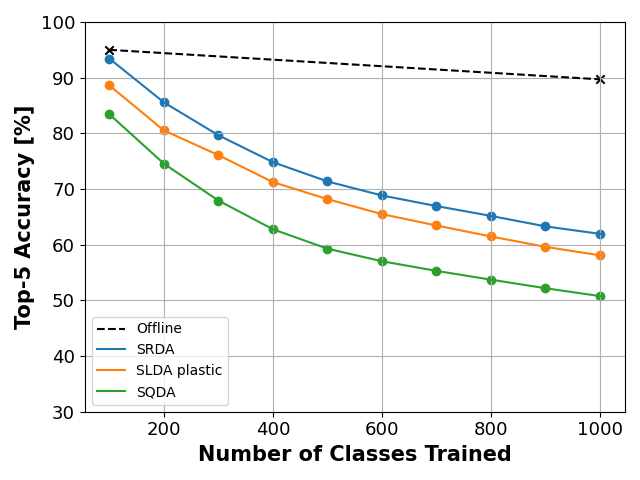}
\end{center}
   \caption{Top-5 Accuracy on ImageNet ILSVRC-2012. We compare our SRDA with $\alpha=0.55$ to SQDA and SLDA with a plastic (non-fixed) covariance matrix.}
\label{fig:long}
\label{fig:onecol}
\end{figure}

For comparison, we use the models and results of \cite{hayes_lifelong_2020} as we follow the same experiment settings. We don’t use models that require task labels as this is not compatible with this more general class incremental learning.

\subsection{Initialization}
As was done in previous works \cite{hayes_lifelong_2020,ferrari_end--end_2018,rebuffi_icarl_2017}, we initialize $F$ and $G$ with a 100 fixed randomly selected classes. We use the same weights for $G$ as \cite{hayes_lifelong_2020} for the first 100. The 900 remaining classes are trained incrementally with a fixed representation for $G$ as mentioned in section \ref{sec:srda}.% For $F$, we initialize the covariance matrices to one and the mean vectors to zero, we then fit the first 100 selected classes in a streaming fashion.
%\begin{itemize}
%    \item In the first ones, we simply start with means of 0, and covariances initialized to 1.
%    \item The second initialization scheme uses Oracle Approximating Shrinkage estimator \cite{chen_shrinkage_2010}. This estimator is supposed to perform better since it is made for high-dimensionality problems with a low number of samples.
%\end{itemize}

\subsection{Results}

\begin{figure}[t]
\begin{center}
   \includegraphics[width=0.95\linewidth]{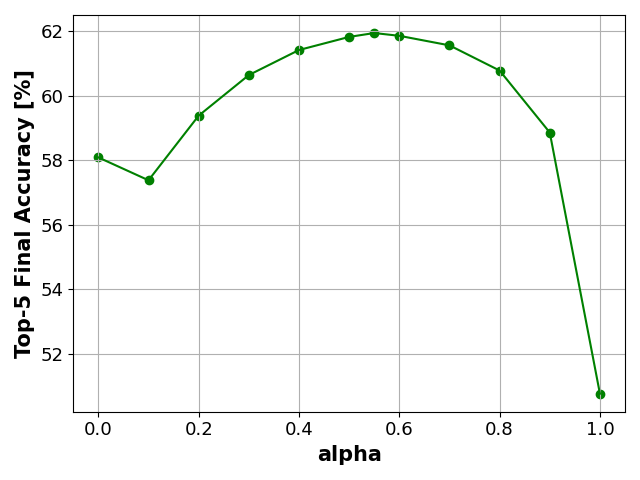}
\end{center}
   \caption{Variations of Top-5 Accuracy on ImageNet ILSVRC-2012 with regard to $\alpha$. An $\alpha=0$ represents a regular SLDA, whereas an $\alpha=1$ represents an SQDA.}
\label{fig:alpha}
\end{figure}

As shown in table \ref{table1}, our method outperforms the streaming state-of-the-art by \textbf{5 \%} and is very close to the offline training of the last layer. SRDA also outperforms iCarl, and End-to-End, which are methods that update the whole model and can iterate multiple times on the data. It should be noted that SQDA, as mentioned earlier, struggles in high-dimensional settings due to the limited per-class availability of data points required for accurate estimation of covariance matrices. To address this limitation, SRDA serves as a corrective measure by leveraging the well-estimated LDA covariance matrix in combination with the estimated class covariance matrices. The figure \ref{fig:alpha} provides visual evidence supporting our findings, with a grid search CV revealing the optimal $\alpha$ value of 0.55 for this experiment. Better results can potentially be achieved by using a more recent backbone, such as EfficientNets \cite{tan_efficientnet_2020}, enabling higher accuracy with a lighter model more adapted to on device-learning.

%It is interesting to note the difference between the accuracy curves of the two initialization schemes \ref{figure}. We notice that with OAS, we are able to reach higher accuracy but struggle to maintain good accuracies for most $\alpha$ values. Whereas, with the first initialization, we are able to have better results than SLDA on most alpha values. Talk about how oas is suited for high-dimensional problems.

\begin{table}[b]
    \label{table1}
    \centering
    \caption{$\Omega_{all}$ accuracy on ImageNet. The results marked with * are taken from \cite{hayes_lifelong_2020} as our experiment follows the same conditions.}
    \begin{tabular}[t]{lcc}%{l>{\raggedright}p{0.3\linewidth}>{\raggedright\arraybackslash}p{0.3\linewidth}}
        \toprule
        Models &Streaming& CLS-IID\\
        \midrule
        \textbf{Output Layer Only:}&&\\
        Fine-Tuning* &Yes& 0.146\\
        ExStream* \cite{hayes_memory_2019}&Yes& 0.569\\
        SLDA \cite{hayes_lifelong_2020}&Yes& 0.752\\
        SQDA (ours) &Yes& 0.677\\%check result
        \textbf{SRDA (ours)} &Yes& \textbf{0.801}\\
        \midrule
        \textbf{Representation Learning:}&&\\
        Fine-Tuning* &Yes& 0.121\\
        iCaRL* \cite{rebuffi_icarl_2017} &No& 0.692\\
        End-to-End* \cite{ferrari_end--end_2018} &No& 0.780\\
        \midrule\midrule
        \textbf{Offline Upper Bounds:}&&\\
        Offline (Last Layer)&No& 0.853\\
        Offline&No& 1.000\\
        \bottomrule
    \end{tabular}
\end{table}

\subsection{Hyperparameter tuning}
Because this method requires the adjustment of a hyperparameter, alpha, one would think that it cannot be readily used out of the box. However, in contrast to regular machine learning hyperparameters, alpha can be modified at the end of training with minimal additional computational costs. This is due to the independent calculation of the two covariance matrices. Consequently, the model can be trained using SRDA and tuned with a quick grid search CV at the end utilizing the validation dataset without retraining. In cases where a validation dataset is unavailable, a potential solution is to maintain a small, class-balanced buffer specifically for hyperparameter tuning, which can be employed at the end of training. This enables this classification technique to be directly used and competitive with other Streaming Learning algorithms.

\subsection{Computation}
\label{sec:computation}

Due to its quadratic complexity, our algorithm takes 12 hours to compute, which is considerably higher than SLDA, which takes 30 minutes on ImageNet. Nonetheless, this is still comparatively manageable compared to other batch learning and streaming algorithms. For example, according to \cite{hayes_lifelong_2020} and our experiences, ExStream takes 64 hours, and iCarl \cite{rebuffi_icarl_2017} 35 hours on the same hardware.

\subsection{Memory usage}
As it is with computational consumption \ref{sec:computation}, SRDA consumes more memory than SLDA as it has to store a covariance per class compared to one covariance matrix in SLDA. For instance, in the case of ImageNet ILSVRC-2012 \cite{russakovsky_imagenet_2015} one needs $(1000\times4\times(512^{2}+512))$ bytes which is equivalent to 1.051 GB. For comparison, SLDA requires 0.001 GB, ExStream requires 0.041GB, and iCarl requires 3.011 GB.

%------------------------------------------------------------------------
\section{Conclusion \& Discussions}
We presented Deep Streaming Regularized Discriminant Analysis, a generative classifier able to adapt to non-iid data streams and outperform existing batch and streaming learning algorithms when paired with a CNN. We outperformed SLDA by 5\%, iCarl by 11\%, and End-to-End by 2\%. This is an impressive result considering that both iCarl and End-to-End update the whole network and should intuitively beat a method only focusing on the last layer.

This method provides better results than SLDA at the cost of computation and memory but remains comparatively manageable compared to other methods. SQDA is better suited for low dimensional and low class counts problems, while SRDA manages to adapt to high dimensional problems with the correct regularization parameter alpha that can be found at the end of training with minimal additional computational costs

For use cases where one would like to combine the speed of SLDA and the performance of SRDA, one can imagine a model where SLDA is used for rapid learning while SRDA slowly trains in the background enabling improved accuracy in the long run.

Finally, this method represents a step forward in the research of Sustainable AI. As presented by \cite{cossu_sustainable_2021}, Continual Learning is a promising candidate for achieving Sustainable AI. This case of Streaming Learning justifies this choice even further as it presents a more realistic application that respects the principles of Sustainable AI, including efficiency, privacy, and robustness. Our deep SRDA has many potential applications, including robotics, edge learning, and human-machine interfaces. It removes the need to store the data as the model can learn on data streams, learning at approximately 28Hz for our experiment on ImageNet. More importantly, it enables on-device Continual Learning, removing the need for retraining and thus saving resources. 

\section*{Appendix}

\noindent The models were trained using these parameters:
\begin{itemize}
    \item \textbf{Offline}: Same parameters as \cite{hayes_lifelong_2020}. SGD for 90 epochs, with $lr=0.1$ with decay at 10 30 and 60 epochs, $momentum=0.9$ and weight decay of $10^{-4}$.
    \item \textbf{iCarl}: Parameters from \cite{rebuffi_icarl_2017}, and stored 20 exemplars per class.
    \item \textbf{ExStream}: Same parameters as offline with 20 exemplars per class.
\end{itemize}

\section*{Acknowledgments}
This work was performed as part of the DREAM project. The DREAM project received funding from the European Union's European Innovation Council (EIC) under grant agreement No 101046451.

{\small
\bibliographystyle{ieee_fullname}
\bibliography{srda}
}

\end{document}